\documentclass[final,3p,times,twocolumn]{elsarticle}

\usepackage{amssymb}
\usepackage{amsmath}

\usepackage[T1]{fontenc}
\usepackage{duffner}
\usepackage{subcaption}

\usepackage{amsfonts}
\usepackage{array}
\usepackage[caption=false,font=normalsize,labelfont=sf,textfont=sf]{subfig}
\usepackage{textcomp}
\usepackage{stfloats}
\usepackage{url}
\usepackage{verbatim}
\usepackage{graphicx}

\usepackage{comment}
\usepackage{booktabs}
\usepackage[ruled]{algorithm2e}
\SetKwComment{Comment}{$\triangleright$\ }{}
\usepackage[separate-uncertainty=true]{siunitx}

\journal{Neural Networks}

\begin{document}

\begin{frontmatter}



\title{Growing Networks with Autonomous Pruning}


\author[liris]{Charles de Lambilly} 
\ead{charles.de-lambilly@etu.ec-lyon.fr}
\author[liris]{Stefan Duffner}
\ead{stefan.duffner@liris.cnrs.fr}

\affiliation[liris]{organization={INSA Lyon, CNRS, Ecole Centrale de Lyon, Universite Claude Bernard Lyon 1, Université Lumière Lyon 2, LIRIS, UMR5205},
            addressline={}, 
            city={Villeurbanne},
            postcode={69621}, 
            country={France}}

\begin{abstract}
This paper introduces Growing Networks with Autonomous Pruning (GNAP) for image classification. Unlike traditional convolutional neural networks, GNAP change their size, as well as the number of parameters they are using, during training, in order to best fit the data while trying to use as few parameters as possible. This is achieved through two complementary mechanisms: growth and pruning. GNAP start with few parameters, but their size is expanded periodically during training to add more expressive power each time the network has converged to a saturation point.
Between these growing phases, model parameters are trained for classification and pruned simultaneously, with complete autonomy by gradient descent. 
Growing phases allow GNAP to improve their classification performance, while autonomous pruning allows them to keep as few parameters as possible. 
Experimental results on several image classification benchmarks show that our approach can train extremely sparse neural networks with high accuracy. For example, on MNIST, we achieved 99.44\% accuracy with as few as 6.2k parameters, while on CIFAR10, we achieved 92.2\% accuracy with 157.8k parameters. 
\end{abstract}



\begin{keyword}
Machine Learning \sep Convolutional Neural Networks \sep Neural Network Pruning \sep Growing Neural Networks
\end{keyword}

\end{frontmatter}


\section{Introduction} \label{Introduction}
Convolutional
Neural Networks (CNNs) have achieved great successes in numerous computer vision tasks such as image classification. While some of these achievements can be attributed to the introduction of new ideas for the training phase of CNNs, many fundamental improvements have come from the search of more effective neural architectures. 
Historically, new architectures were developed by researchers through a difficult, time-consuming, trial-and-error method. 
However, in recent years, there has been a growing interest in ways to automate and optimise this architecture design phase.

A popular approach is called Neural Architecture Search (NAS), and it has achieved great progress recently, spawning over 1000 papers in the last four years \cite{NAS}. NAS methods can be distinguished through their search space, search strategy and performance evaluation strategy. The search space is the realm of possible architecture NAS can select. The search strategy is the strategy NAS uses to explore the search space. Finally, the performance evaluation strategy is the way NAS evaluate architectures coming under its consideration. NAS has been shown to be able to surpass human designers in many tasks. However, despite various improvements in its efficiency they remain computationally expensive because many models with different hyperparameters need to be trained.

Another common way of automating the architecture design phase is pruning. By training a highly over-parameterised neural network compared to what would theoretically be necessary to adequately model the data, and then pruning its weights, it is possible to keep the performance of the large network while only keeping few of its weights. However, training the very large network is still a computationally expensive task, and it can also easily lead to overfitting.
Moreover, the design and sizing of the architecture of the initial network is still a manual process that may be sub-optimal. 
For example, the choice of the number of layers is critical and is usually not reduced by pruning.

Alternatively, it is also possible to automate the design phase with a growing network. The idea is to do the exact opposite of pruning, i.e.\ train a small network, and then periodically add neurons to it to help it model better the data. There are three different questions that every growing network architecture must address: when to add neurons, where to add neurons, and how to initialise the weights of the new neurons \cite{GradMax}. 
Many existing such architectures have taken inspiration from real-world neural growth of biological nervous systems. However, steering the growth of the neural network in a way that makes sure that the added neurons are strictly necessary remains a challenge.
Therefore, the risk is that the resulting network is larger than their fixed-size counterparts with comparable classification accuracy as each growing step may add some redundancy in the model. 

To address these issues, in this paper, we use growing and pruning in a combined optimization process that changes the topology and size of the neural network during training while minimising its classification loss.
We introduce Growing Networks with Autonomous Pruning (GNAP), a novel way of using pruning to keep a growing network as small as possible while still letting the growing phases improve classification performance. 
The overall approach consists in jointly training the weights and the topology of a specific CNN architecture encouraging sparse networks while repeatedly growing its structure whenever the sparsity-inducing pruning approaches saturation. 

Our contributions are the following:
\begin{itemize}
\item a new approach for training neural networks adapting dynamically their size and topology during training by means of an alternating growing and pruning strategy and
\item a specific dynamic and highly flexible CNN architecture 
that allows sparse connections from any layer to any succeeding layer.
\end{itemize}
We experimentally show on three public image classification benchmarks that our proposed approach leads to extremely sparse CNN with a high classification accuracy compared to the state of the art in sparse and efficient CNN models. 
Finally, we also show that our method can easily be extended to structured pruning, where instead of automatically removing single weights, we prune entire convolution kernels (\ie 2D matrices).

\newpage
\section{Related Work} \label{RelatedWork}

Several methods have been proposed in the literature to produce very sparse neural networks. 
Many of them start from dense architectures and apply pruning in such a way that very high pruning rates ($>90\%$) can be obtained, \eg either before training~\cite{lee2018} via a weights sensitivity analysis or based on the lottery ticket hypothesis~\cite{frankle2018, wang2019, su2020, frankle2020} that effectively identifies sub-networks from a trained dense network and retrains these sub-networks. 
Another family of methods is based on NAS, like MCUNet~\cite{lin2020_mcunet}, where sparse networks are obtained by formulating an optimisation problem searching for the best configuration (hyperparameters, network architecture etc.) based on a pre-defined search space.
In the following, we will focus on methods based on sparsity-inducing pruning and growing neural networks.

\noindent {\it Sparsity-inducing training:}\\[1mm]
Sparsity in deep neural networks can be achieved through different techniques. 
One of them is to encourage sparse connectivity or topologies of the neural network during training by introducing an additional term in the loss function, e.g.\ the $\ell_1$ loss~\cite{schmidt2007}. Pruning is then performed by removing the components that have a value close to zero.
This term can be defined for fully-connected layers but also for feature maps or convolution kernels/filters in CNNs~\cite{liu2017,yang2019,sun2019,kumar2020}.

In ART~\cite{Glandorf_2023_ICCV}, the sparsity-inducing regularisation term is \emph{adaptive} during training and allows to gradually shift from exploration and exploitation in the optimisation process. 

Another approach proposed by~\cite{mocanu2018}, called Sparse Evolutionary Training (SET), is to repeatedly prune a certain fraction of weights in each layer during training and replace them by new random values. 
This has then been extended in~\cite{mostafa2019} by a dynamic sparse reparameterisation (DSR) algorithm, where this pruning rate and allocation of weights is dynamically set across layers thus leading to more sparse architectures.

The method introduced in \cite{SGDPruning} achieves neural network pruning during training through gradient descent, which inspired the approach proposed in this paper. The main idea is to link every weight of the neural network to a gate with learnable parameters, and define the loss function as the sum of a prediction loss and a sparsity loss. During each forward pass, every gate can be opened or closed, with a probability determined by the learnable parameters of every gate. If a gate is closed, the corresponding weight is set to zero for that forward pass, otherwise it is kept as it is. The sparsity loss is determined by the number of opened gates, meaning that in order to have the lowest loss possible, the model must try to fit the data as well as possible while using as few parameters as possible. 
For further details, refer to subsection \ref{AutonomousPruning}, which details both the original idea~\cite{SGDPruning} and the adaptations made for GNAP. 
A similar idea, called "SimpleNet", was also presented in \cite{SimpleNet}, with some differences, the most notable one being that the approach is instead deterministic.

\vspace{2mm}
\noindent {\it Growing neural networks:}\\[1mm]
Other research works propose to start training a small network (sometimes called seed network) and iteratively add more neurons, feature maps or layers. 
The advantage with respect to pruning is that we do not need to set an "upper bound" size of the overparameterised model in advance, which additionally may lead to overfitting.
However, an important issue with approaches that grow networks is to prevent the models from adding redundancy while still allowing them to improve their performance. 

To address these issues, the Splitting Steepest Descent approach~\cite{liu2019}, for example, splits existing neurons into two during training by formulating the splitting procedure as an optimisation problem which determines when and where to split in order to increase the model's performance. 
An extension, the firefly neural architecture descent algorithm~\cite{wu2020} grows a larger network from a small one by iteratively defining the possible growth as a functional neighborhood and a greedy selection procedure. 
Growing is performed by either splitting a neuron, adding a new neuron or adding a new layer, thus allowing to increase both the width and the depth of the architecture.

In the GradMax approach~\cite{GradMax}, a new growing strategy is introduced that starts from a seed network architecture (MLP or CNN-based) and at every added neuron tries to minimise its impact on the already learnt weights while maximising the learning dynamics for the subsequent training of the grown network. 
This is achieved by initialising the new incoming weights to zero and maximising the gradient of the new outgoing weights using Singular Value Decomposition (SVD).

In~\cite{yuan2023}, the authors propose a new growing algorithm that initialises the new weights and rescales the old ones according to a variance transfer strategy. 
Also local learning rates are set specifically for the added subnetworks in order to prevent divergence.

Recently, \cite{TINY} introduced a way to measure expressivity bottlenecks in neural networks, and reduce them optimally. The "expressivity bottleneck" of a layer $l$ is defined as the mean of the difference between how every datapoint $x_i$ would like the pre-activation output $a_l$ of the layer to update in order to minimise the loss due to $x_i$, and how $a_l$ will actually update due to the contradictory needs of the datapoints. Thanks to this measure, the layers lacking expressive power can be identified and improved, by adding to them neurons with weights that allows the update of $a_l$ to more closely match the desired update of every datapoint.

\vspace{2mm}
\noindent {\it Combined growing and pruning:}\\[1mm]
Combining growing networks with pruning takes advantage of both approaches, and a few methods have been proposed in the literature recently.
For example, the sparse momentum algorithm from~\cite{dettmers2019} (Sparse Networks from Scratch, SNFS) is similar to the DLR method in~\cite{mostafa2019}. Here the pruned weights are redistributed to layers that have larger average gradients, and new weights are grown between neurons with the highest momentum magnitude. 
An alternative approach called RigL is presented in~\cite{evci2020}, where the sparsity constraints across layers are fixed in the beginning and connections with the highest gradients are grown. 

To improve the efficiency and performance with Generative Adversarial Networks (GAN), Self-Growing and Pruning Generative Adversarial Networks (SP-GAN) have been introduced by \cite{SP-GAN}. 
The general idea is quite similar to ours, starting from two small networks (for the generator and discriminator), the "seed networks", that are first trained on their own. 
Once the training of these seed networks is completed, they both enter "self-growing phase", were the convolutional filter of all convolution layers is replicated, after which the networks are trained again. This process can then be repeated until the networks reach satisfactory performance. 
They then enter the final phase, the "pruning phase". During that phase, for every layer, the Euclidean distance between every pair of feature map is computed. The convolutional filters corresponding to the k-smallest distance can then be pruned, allowing the networks to reduce their size while still keeping their performances by removing redundant filters.

Our GNAP approach also combines pruning with growing. However, it is fundamentally different from these methods in the following ways. 
Instead of a separate pruning step at regular intervals based on some criteria (e.g.\ weight magnitude) we adopt a gating-based pruning strategy~\cite{SGDPruning} where weights and network topology is optimised jointly, and we adapted it to our specific CNN structure. 
The growing steps however are done iteratively whenever the pruning and weight optimisation converges to a solution. 
In this way, the new neurons or convolution filters may be trained in a more complementary way and only if/when the algorithm considers it necessary.
Finally, our initially very densely connected dynamic CNN architecture gives the model many degrees of freedom and allows to produce extremely sparse networks at convergence.

\section{Growing Networks with Autonomous Pruning}

\subsection{Architecture}

\begin{figure*}
  \centering
  \includegraphics[width=\textwidth]{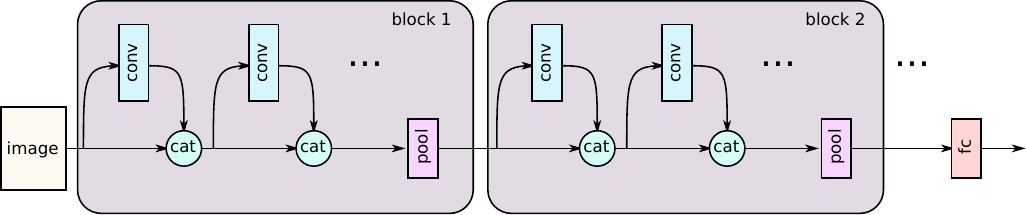}
  \caption{The overall network architecture of GNAP (\emph{conv}: convolution layer, \emph{pool}: pooling layer, \emph{cat}: feature map concatenation, \emph{fc}: fully-connected layer). Each layer may be (sparsely) connected to all preceding layers. Each block may contain one or several convolution layers with the same feature map size. During training, structured (kernel) pruning or unstructured pruning is performed and layers are grown and added repeatedly.}
  \label{fig:gnap_architecture}
\end{figure*}

The philosophy behind GNAP is to give the network as many architecture choices as possible, and let it choose the ones that best fit the task at hand. 
For this reason, GNAP took inspiration from the architecture of Densely Connected Convolutional Networks (DenseNets) \cite{DenseNet}, which connects each convolutional layer to every subsequent layers of the model. This makes sure that no knowledge is lost during the forward pass, since every previously generated feature map is available at each layer. 
However, since pooling layers are an essential part of CNNs, and feature maps of different sizes can not be concatenated together, DenseNets are organised into dense blocks, where all feature maps have the same size, with a pooling layer between dense blocks to down-sample the layers. 
GNAP work the same way, retaining the idea of dense blocks separated by pooling layers. 
Figure~\ref{fig:gnap_architecture} illustrates the overall architecture of GNAP.
DenseNets also use additional ideas like compression and bottleneck to prevent an excessive growth in the number of feature maps, but GNAP do not need them, since the autonomous pruning process presented in subsection \ref{AutonomousPruning} allows by itself to keep the number of feature maps small. 
In contrast to DenseNets, GNAP are growing neural networks, i.e.\ they dynamically change their architecture during training. 
In our experiments, the number of blocks $B$ (and thus the number of pooling layers) is fixed, but the width of each layer (i.e.\ the number of channels or neurons) and the number of convolution layers in each block is varying.
We keep a single fully-connected (fc) layer at the end.

\subsection{Autonomous Pruning} \label{AutonomousPruning}

As stated in section \ref{RelatedWork}, the autonomous pruning process of GNAP is based on Neural Network Pruning by Gradient Descent, previously described in \cite{SGDPruning}. 
In order to achieve pruning by gradient descent, two "gating" parameters (one for \emph{open} and one for \emph{closed}) are created for every weight of the model. Then, during each forward pass, for every gate, these gating parameters are used as the parameters of a Gumbel-Softmax distribution \cite{GumbelSoftmax}, from which a gating variable (either zero or one) is sampled, and then multiplied with the corresponding weight. Since sampling from a Gumbel-Softmax distribution is differentiable, gradient descent can then update both the weights of the models as well as the gating parameters simultaneously. Weights can therefore be "pruned" in a differentiable way by gradually decreasing, during optimization, the probability that their corresponding gate is opened. In order to give the model an incentive to prune weights, the loss is defined as: 
\begin{equation} 
    L=L_{pred}+\alpha \cdot \left|\frac{1}{N}\sum _{i=0}^{N-1}g_i-D_{target} \right| \; ,
\label{eq:original-loss}
\end{equation}
\ie the sum of a prediction loss $L_{pred}$ -- here the classical cross-entropy loss for classification -- and a sparsity loss.
The scalar $\alpha$ is the weight of the sparsity loss, $N$ is the number of weights, $g_i$ is the value of the gating variable of weight $i$ and $D_{target}$ is the desired weight density of the network, i.e.\ the proportion of weights with open gates. 
Thanks to this loss, the model is encouraged to both minimise the prediction loss and be as close as possible to the target weight density. 

In GNAP, however, autonomous pruning works slightly differently. The main reason is that the size of the network is not fixed, meaning a weight density is not a relevant metric for the sparsity loss. 
Therefore, the number of gates is used instead, which is equivalent to not multiplying $\sum _{i=0}^{N-1}g_i$ by $\frac{1}{N}$. 
Another difference is that, for GNAP, $D_{target}$ is always equal to zero, since we want to use as few weights as possible.
Thus, the overall loss we propose is the following:
\begin{equation} 
    L=L_{pred}+\alpha \cdot ||g||_1 \; .
\label{eq:loss}
\end{equation}

The biggest difference, though, is that while method of~\cite{SGDPruning} only allows to automate unstructured pruning (since only individual weights could be pruned), GNAP also allows for structured pruning. 
Like the original paper, GNAP can have a gate for every weight, but we can also chose to use a gate for every convolutional kernel. This allows to do both structured and unstructured pruning, depending on the needs of the application or the underlying hardware at inference time. 
Thus the implementation of these different types of pruning is rather simple: gating variables are still sampled from Gumbel-Softmax distributions, but instead of being multiplied with a single weight, they are multiplied with an entire convolutional kernel.

\subsection{Growing Phases}
\label{sec:growing}
As stated in section \ref{Introduction}, every growing network algorithm needs to determine when to add neurons, where to add neurons, and how to initialise the weights of the new neurons. We will now present how GNAP answer all of these questions.
Note that, in the following, we employ the term \emph{neuron} irrespective of the type of layer: convolution layer or fully-connected layer. 
For a convolution layer, this corresponds to a convolution filter or a feature map/channel.

Ideally, we should only add neurons when necessary, i.e.\ when the network cannot further improve its performance with the given architecture. This means waiting for the network to converge to a stable state. 
In GNAP, though, gradient descent optimises both prediction loss and topology, and we have to wait for the two loss terms to converge. 
In practice however, we observed that the prediction loss always converges before the sparsity loss. 
Therefore, the criterion we use for deciding when to add neurons only takes topology into account. 
We define a criterion based on the relative change in open gates of the model:
\begin{equation} \label{eq2}
    \frac{n_{mean}-n_{gates}}{n_{mean}}<\theta
\end{equation}
where $n_{gates}$ is the current number of open gates, $n_{mean}$ is a running mean of the value of $n_{gates}$ across the last $M$ epochs.  
After each epoch, this criterion is calculated, and when it is smaller than the threshold $\theta$, a growing phase is launched.
Training is performed for $E$ epochs in total. However, we allow growing phases only until epoch $E'$ (in our experiments $30\%$ of $E$). 
In the second phase after $E'$, this gives the topology of the network more time to converge, and the algorithm usually finds smaller models with equivalent classification accuracy.

The choice we made regarding the question as to where to add new neurons is to not make any choice and add neurons everywhere: $C$ new neurons are added to every layer, and a new layer is added at the end of every block. 
As with the initial architecture, all new neurons are connected to all preceding neurons (including the new ones).
This choice is reasonable because afterwards, the autonomous pruning process will remove redundant neurons and connections and only keep useful ones. 
Furthermore, adding neurons this way is consistent with the philosophy of GNAP: give the network as many possibilities as possible, and let it choose for itself what it finds most useful among those possibilities. 

As for how to set the initial weights of the newly created neurons, there are two things to consider. 
The first is how to connect the new neurons amongst themselves. 
The second is how to connect the new neurons to the existing ones. 
In the first case, we chose to set the weights randomly from a Gaussian distribution $\mathcal{N}(0,\sigma)$ with $\sigma = 1/n_i$ and $n_i$ being the number of inputs to the neuron.
Concerning the second case, the weights connecting the outputs of new neurons to the inputs of existing ones, risks destabilising the network by feeding "garbage" data to trained neurons.
In initial experiments, we therefore set those weights to zero. What we discovered, though, is that this strategy ended up making the model dismiss all the neurons added by the growing phases, since they increased the sparsity loss without helping its existing, trained weights. 
Even if the sparsity loss was decreased after a growing phase, the newly-created weights could not compete with the already-trained weights. 
Therefore, allowing the new neurons to feed "garbage" data to the existing ones fixed this issue, since the model had to train the new weights to make them give useful data to the old weights.
We decided to use the same gaussian distribution as described above, $\mathcal{N}(0, \sigma)$, for this initialisation of the weights connecting the ouputs of new neurons to the inputs of old neurons.

Regarding the weights connecting the outputs of old neurons to the inputs of new neurons, we could simply use $\mathcal{N}(0, \sigma)$ again. However, we instead
propose the following more effective strategy. 
For a given layer $l$, let $W_l = \{W(i,j)\} \in \mathbb{R}^{m \times n}$ be the (centered) weight matrix of the $n$ existing neurons in the layer (column $n$ being the weights of neuron $m$). 
Equivalently, for convolution layers, the columns are formed by vectorising the convolution filters of each channel.
To obtain the weight vector of a new neuron $w$, we first calculate the Singular Value Decomposition (SVD) of $W_l$:
\begin{equation}
USV = \mathrm{SVD}(W_l) \; ,
\label{eq:svd}
\end{equation}
and then compute a vector $w' \in \mathbb{R}^m$ that is orthogonal to the basis $V \in \mathbb{R}^{m \times m}$. 
To this end, we generate a random vector $r  \in \mathbb{R}^m$ independently from $\mathcal{N}(0,1)$ and concatenate it as a new column to $V$:
\begin{equation}
  V' = (V r) \; .
\end{equation}
Then, we obtain $w'$ by solving the least-square problem:
\begin{equation}
  w V' = c \; ,
\end{equation}
where $c = (0, 0, \ldots, 0, 1) \in \mathbb{R}^{m+1}$. 
Finally, we assign $w$ the absolute values of $w$ with random signs $s \in \{-1, +1\}^m$:
\begin{equation}
  w = s \abs(w') \; .
\label{eq:randsign}
\end{equation}
We empirically found that this balancing of positive and negative weights is beneficial for the performance.
If $n<m$, i.e.\ the rank of $W_l$ is smaller than the dimension of weight vectors, instead of using this algorithm, we simply create a random vector $w$ with $w(i) \sim \mathcal{N}(0, \sigma)$.
Finally, the vector $w$ is concatenated to the matrix $W_l$ and "uncentered".

\subsection{Training procedure}
Algorithm~\ref{alg:gnap-train} summarises the overall training procedure.
\begin{algorithm}[t]
\footnotesize
  \DontPrintSemicolon
    \KwIn{Loss $L$, weights $W$, number of epochs $E$, epoch from when to stop growing $E'$, number of new neurons $C$}
    
    \ForEach{$e \in (1..E)$}{
      GradientDescent(L, W) \Comment*[r]{minimise Eq.~\ref{eq:loss}} 
      \If{$\frac{n_{mean}-n_{gates}}{n_{mean}}<\theta$ and $E<E'$}{
        \tcp{Growing phase}
        \ForEach{$b \in (1..B)$ \Comment*[r]{loop over blocks}}
        {
          \ForEach{$l \in (1..K_b)$ \Comment*[r]{loop over layers in block $b$}}
          {
            Retrieve existing weights: $W_l \in \mathbb{R}^{m \times n}$ \;
            \For{1..C}{
            \uIf{$n < m$}
            {
              $w(i) \sim \mathcal{N}(0, \sigma) \quad \forall i$ \Comment*[r]{Gaussian weights}
            }
            \Else
            {
              $w \leftarrow \mathrm{orthogonal}(W_l)$ \Comment*[l]{orthogonal weights (Eq.\ref{eq:svd}-\ref{eq:randsign})}
            }
            Concatenate $W_l$ and column vector $w$
            } 
            Add random weights with values from $\mathcal{N}(0, \sigma)$ to neurons from succeeding layers
          }  
          Add new layer to block $b$ with weight values from $\mathcal{N}(0, \sigma)$\\
          Add random weights with values from $\mathcal{N}(0, \sigma)$ to neurons from succeeding layers
        }  
      }
      Compute validation error and number of links for early stopping
    }
 \caption{GNAP training algorithm.}
 \label{alg:gnap-train}
\end{algorithm}

The GNAP network is trained for $E$ epochs, and we compute the accuracy and number of links (i.e.\ open gates) after each training epoch. 
We perform early stopping in order to save the model with the best performance on the validation set. 
However, to be able to retain models that have a good tradeoff between accuracy and size, we modified the classical early stopping criterion based on the accuracy only to also include the number of active links.
Concretely, we save a model if its accuracy is higher or if its number of links is lower with an equivalent accuracy. 
And we consider two accuracies equivalent if their difference is below a threshold $\theta_a$. 
This allows us to obtain models that are much sparser while still having an acceptable classification accuracy.

\subsection{Inference}

The inference is done in a classical feed-forward way. The only difference with respect to the forward pass during training is the gating weight. 
Instead of the stochastic method, where each gate is sampled from the respective Gumbel-Softmax distribution, a deterministic \textit{softmax} function is used, and the gate is opened if the probability (softmax) is above the trivial threshold $\gamma=0.5$ and closed otherwise.

\section{Experiments}
We evaluated our approach on three image classification benchmarks: MNIST, CIFAR10 and CIFAR100. 
We used 10\% of the training sets for validation and early stopping and 90\% for training and report the average classification accuracies for the respective test sets (with 10 runs).  

\subsection{Hyperparameters}
For training we used the following data augmentation operations on the images: horizontal flip (50\%), rotation (max. 15°), affine transformations (shear 10, translate 0.1), crop (0.8..1.0) and cut out.

All models were trained using adaptive moment estimation (Adam), with default parameters and a weight decay of $10^{-6}$. 
The batch size was set to 256 and the learning rate to 0.01. 
For GNAP-specific hyperparameters, the sparsity loss weight $\alpha$ was set to $0.5 \times 10^{-7}$, the growing phase threshold $\theta$ was set to 0.05 and the running mean of the number of opened gates $n_{mean}$ was computed over the last 10 values. 
The tolerated accuracy decrease $\theta_a$ is set to 0.5\% for CIFAR10 and CIFAR100 and to 0.25\% for MNIST.
In our experiments, we performed a fixed number of growing phases: $12$. After the final phase, we perform a refinement phase of 200 epochs with an increased weight $\alpha=0.25 \times 10^{-5}$ for the topological loss term. This allows the algorithm to converge to even sparser networks at the end of training.

Since MNIST and CIFAR images are relatively small (respectively 28x28 and 32x32 pixels), we only used two dense blocks for MNIST and three for CIFAR. 
For MNIST, our initial architecture only had 1 convolution layer per block with size 10. 
For CIFAR10 and CIFAR100, we used three layers with sizes 64 for the first block, three layers of size 128 for the second block and three layers of size 256 for the third block.
Table~\ref{tab:architectures} lists all the initial architectures that have been used in our experiments in order to achieve different accuracy-complexity trade-offs.
\begin{table}
\centering
\caption{Initial architectures of the different tested models obtaining different accuracy-complexity trade-offs. The numbers indicate the size of the layers (i.e.~the number of neurons/channels).}
\label{tab:architectures}
\footnotesize 
\begin{tabular}{l@{\hspace{0.5em}}c@{\hspace{0.6em}}c@{\hspace{0.6em}}c@{\hspace{0.5em}}c}
\toprule
Model & block 1 & block 2 & block 3 & FC \\
\midrule
GNAP (MNIST) & 10 & 10 & & 10 \\
GNAP (CF10) & 64, 64, 64 & 128, 128, 128 & 256, 256, 256 & 10 \\
GNAP (CF100) & 64, 64, 64 & 128, 128, 128 & 256, 256, 256 & 100 \\
NAP (m1) (MNIST) & 10 & 10 & & 10 \\
NAP (m2) (MNIST) & 100 & 100 & & 10 \\
NAP (m3) (CF10) & 64 & 128 & & 10 \\
NAP (m4) (CF10) & 64, 64 & 128, 128 & & 10 \\
NAP (m5) (CF10) & 64, 64, 64 & 128, 128, 128 & & 10 \\
NAP (m6) (CF10) & 64, 64, 64 & 128, 128, 128 & 256, 256, 256 & 10 \\
NAP (m7) (CF100) & 64, 64, 64 & 128, 128, 128 & & 100 \\
NAP (m8) (CF100) & 64, 64, 64 & 128, 128, 128 & 256, 256, 256 & 100 \\
\bottomrule
\end{tabular}
\end{table}
In theory, GNAP can start from small architectures and increase them to very large ones by numerous growing phases. 
In practice however, it is preferable to use a more complex initial architecture if one knows that the data is more difficult (e.g.\ CIFAR100 compared to CIFAR10) because, otherwise, the total number of epochs (and thus training time) must be considerably increased.
An alternative would be to increase $C$, the number of new neurons, added at each growing phase.
We have not yet studied the impact of these different parameters on the training dynamics, convergence and final performance.

\subsection{Results}

Tables~\ref{tab:mnist}-\ref{tab:cifar100} show the results for the datasets MNIST, CIFAR10 and CIFAR100.
For comparison, we evaluated a version of our GNAP algorithm without the growing stage, called "NAP" and with different initial architectures (cf.\ Table~\ref{tab:architectures}).
Unstructured and structured pruning types were tested: denoted by "-U" and "-S" respectively.
And for the growing networks GNAP, we compared our proposed orthogonal weight initialisation (see Section~\ref{sec:growing}) with a standard initialisation from a normal distribution.

\begin{table*}[htbp]
\centering
\caption{Performance on MNIST.}
\label{tab:mnist}
 \begin{tabular}{l@{\hspace{1em}}r@{\hspace{0.5em}}l@{\hspace{1em}}r@{\hspace{0.5em}}l@{\hspace{1em}}r@{\hspace{0.5em}}l@{\hspace{1em}}r@{\hspace{0.5em}}l}
\toprule
Method & \multicolumn{2}{c}{Accuracy} & \multicolumn{2}{c}{Links} & \multicolumn{2}{c}{Weights} \\
\midrule
NAP-U (m1) & 0.9875 & $\pm$ 0.0008 & 5762 & $\pm$ 437 & 5762 & $\pm$ 437 \\
NAP-U (m2) & 0.9938 & $\pm$ 0.0008 & 9264 & $\pm$ 2202 & 9264 & $\pm$ 2202 \\
GNAP-U (rand) & 0.9943 & $\pm$ 0.0004 & 6154 & $\pm$ 291 & 6154 & $\pm$ 291 \\
GNAP-U (ortho) & 0.9944 & $\pm$ 0.0004 & 6234 & $\pm$ 203 & 6234 & $\pm$ 203 \\
\midrule
NAP-S (m1) & 0.9823 & $\pm$ 0.0023 & 5409 & $\pm$ 1388 & 5928 & $\pm$ 1482 \\
NAP-S (m2) & 0.9923 & $\pm$ 0.0012 & 21003 & $\pm$ 10068 & 28533 & $\pm$ 11678 \\
GNAP-S (rand) & 0.9946 & $\pm$ 0.0006 & 5221 & $\pm$ 1151 & 14504 & $\pm$ 2399 \\
  GNAP-S (ortho) & \textbf{0.9947} & $\pm$ 0.0008 & \textbf{4971} & $\pm$ 629 & \textbf{13784} & $\pm$ 1916 \\
\bottomrule
\end{tabular}
\end{table*}

\begin{table*}[htbp]
\centering
\caption{Performance on CIFAR10.}
\label{tab:cifar10}
\begin{tabular}{l@{\hspace{0.5em}}r@{\hspace{0.5em}}l@{\hspace{0.5em}}r@{\hspace{0.5em}}l@{\hspace{0.0em}}r@{\hspace{0.5em}}l}
\toprule
Method & \multicolumn{2}{c}{Accuracy} & \multicolumn{2}{c}{Links} & \multicolumn{2}{c}{Weights} \\
\midrule
NAP-U (m4) & 0.8763 & $\pm$ 0.0055 & 235655 & $\pm$ 67951 & 235655 & $\pm$ 67951 \\
NAP-U (m5) & 0.8849 & $\pm$ 0.0107 & 245168 & $\pm$ 68530 & 245168 & $\pm$ 68530 \\
NAP-U (m6) & 0.9162 & $\pm$ 0.0024 & 696837 & $\pm$ 116179 & 696837 & $\pm$ 116179 \\
GNAP-U (rand) & 0.9245 & $\pm$ 0.0046 & 179979 & $\pm$ 26074 & 179979 & $\pm$ 26074 \\
GNAP-U (ortho) & 0.9222 & $\pm$ 0.0041 & \textbf{157765} & $\pm$ 18620 & \textbf{157765} & $\pm$ 18620 \\
\midrule
NAP-S (m4) & 0.8802 & $\pm$ 0.0049 & 99097 & $\pm$ 38844 & 340501 & $\pm$ 65564 \\
NAP-S (m5) & 0.8912 & $\pm$ 0.0051 & 143666 & $\pm$ 29348 & 521724 & $\pm$ 88896 \\
NAP-S (m6) & 0.9155 & $\pm$ 0.0020 & 311167 & $\pm$ 28118 & 2244999 & $\pm$ 228665 \\
GNAP-S (rand) & 0.9331 & $\pm$ 0.0022 & 88365 & $\pm$ 27679 & 681377 & $\pm$ 200233 \\
GNAP-S (ortho) & 0.9305 & $\pm$ 0.0048 & {88227} & $\pm$ 31551 & {668941} & $\pm$ 220206 \\
\bottomrule
\end{tabular}
\end{table*}

\begin{table*}[htbp]
\centering
\caption{Performance on CIFAR100.}
\label{tab:cifar100}
\begin{tabular}{l@{\hspace{0.5em}}r@{\hspace{0.5em}}l@{\hspace{0.5em}}r@{\hspace{0.5em}}l@{\hspace{0.5em}}r@{\hspace{0.5em}}l}
\toprule
Method & \multicolumn{2}{c}{Accuracy} & \multicolumn{2}{c}{Links} & \multicolumn{2}{c}{Weights} \\
\midrule
NAP-U (m7) & 0.5583 & $\pm$ 0.0173 & 393330 & $\pm$ 121100 & 393330 & $\pm$ 121100 \\
NAP-U (m8) & 0.6408 & $\pm$ 0.0158 & 410827 & $\pm$ 118481 & 410827 & $\pm$ 118481 \\
GNAP-U (rand) & 0.6692 & $\pm$ 0.0106 & 207289 & $\pm$ 59319 & 207289 & $\pm$ 59319 \\
GNAP-U (ortho) & \textbf{0.6656} & $\pm$ 0.0086 & \textbf{179857} & $\pm$ 28182 & \textbf{179857} & $\pm$ 28182 \\
\midrule
NAP-S (m7) & 0.5085 & $\pm$ 0.0124 & 231668 & $\pm$ 3532 & 247850 & $\pm$ 4221 \\
NAP-S (m8) & 0.5965 & $\pm$ 0.0141 & 235029 & $\pm$ 34575 & 461149 & $\pm$ 142521 \\
GNAP-S (rand) & 0.6812 & $\pm$ 0.0067 & 135354 & $\pm$ 17064 & 309983 & $\pm$ 38011 \\
GNAP-S (ortho) & \textbf{0.6784} & $\pm$ 0.0055 & \textbf{125225} & $\pm$ 13379 & \textbf{275970} & $\pm$ 28591 \\
\bottomrule
\end{tabular}
\end{table*}

Here, our aim is to have a model with fewer parameters without loosing too much of accuracy.
We can see that training with growing phases produces models with much fewer parameters for all the three datasets and with structured and unstructured pruning. 
The structured pruning models are slightly bigger than the unstructured ones.
This is because the sparsity-inducing training loss has more difficulties in removing complete convolution kernels as opposed to isolated weights.
The training with our proposed orthogonal weight initialisation generally leads to models that are sparser with equivalent accuracies (except for MNIST with unstructured pruning where the sparsity is roughly the same between the two). 

We also compared our results to those of state-of-the-art approaches producing very sparse neural network models for CIFAR10 and CIFAR100 (see Tables~\ref{tab:sota_cifar10} and~\ref{tab:sota_cifar100}). 
These methods are mostly based on pruning, except MCUNet~\cite{lin2020_mcunet} which is based on NAS and SimpleNet~\cite{SimpleNet} which is a "handcrafted" static CNN architecture.
In the literature, there are many more results of neural network compression algorithms. 
However, most of them are not comparable because they cope with much bigger models (a simple ResNet18 has already around 20M parameters) and target pruning/sparsity rates of only around 30-40\% (instead of around 98-99.5\%). 
Also, they aim for higher accuracies and often use models that are pre-trained on ImageNet or similar.
\begin{table}[htbp]
\centering
\caption{Comparison with state-of-the-art sparse CNN on CIFAR10. Our approach outperforms most methods combining the criteria of accuracy and number of weights without the dependence on a pre-defined initial dense network.}
\label{tab:sota_cifar10}
\footnotesize
\begin{tabular}{l@{\hspace{0.5em}}r@{\hspace{0.5em}}l@{\hspace{0.5em}}r@{\hspace{0.5em}}l@{\hspace{0.5em}}c}
\toprule
Method & \multicolumn{2}{c}{Accuracy} & \multicolumn{2}{c}{Weights} & Base architecture \\
\midrule
SimpleNet~\cite{SimpleNet} & \multicolumn{2}{c}{0.9198}  & \multicolumn{2}{c}{310000}  & \\
SNIP~\cite{lee2018} & 0.9272 & $\pm$ 0.0018 & \multicolumn{2}{c}{186013} & ResNet32 \\
SNIP~\cite{lee2018} & 0.7612 & $\pm$ 0.2196 & \multicolumn{2}{c}{410968} & VGG19 \\
GraSP~\cite{wang2019} & 0.9286 & $\pm$ 0.0019 & \multicolumn{2}{c}{186013} & ResNet32 \\
GraSP~\cite{wang2019} & 0.9216 & $\pm$ 0.0014 & \multicolumn{2}{c}{410968} & VGG19 \\
SRatio~\cite{su2020} & 0.9185 & $\pm$ 0.0016 & \multicolumn{2}{c}{93006} & ResNet32 \\
SRatio~\cite{su2020} & 0.9233 & $\pm$ 0.0024 & \multicolumn{2}{c}{410968} & VGG19 \\
LTH~\cite{frankle2018} & 0.9268 & $\pm$ 0.0032 & \multicolumn{2}{c}{186013} & ResNet32 \\
LTH~\cite{frankle2018} & 0.4117 & $\pm$ 0.4276 & \multicolumn{2}{c}{410968} & VGG19 \\
IMP~\cite{frankle2020} & 0.9135 & $\pm$ 0.0018 & \multicolumn{2}{c}{37203} & ResNet32 \\
IMP~\cite{frankle2020} & 0.9128 & $\pm$ 0.0029 & \multicolumn{2}{c}{205484} & VGG19 \\
RigL~\cite{evci2020} & \textbf{0.9307} & $\pm$ 0.0022 & \multicolumn{2}{c}{\textbf{93006}} & ResNet32 \\
RigL~\cite{evci2020}  & 0.9141 & $\pm$ 0.0015 & \multicolumn{2}{c}{410968} & VGG19 \\
ART~\cite{Glandorf_2023_ICCV} & \textbf{0.9269} & $\pm$ 0.0022 & \multicolumn{2}{c}{\textbf{37203}} & ResNet32 \\
ART~\cite{Glandorf_2023_ICCV} & \textbf{0.9291} & $\pm$ 0.0010 & \multicolumn{2}{c}{\textbf{102742}} & VGG19 \\
MCUNet~\cite{lin2020_mcunet} & \multicolumn{2}{c}{0.8970} & \multicolumn{2}{c}{210100} &  \\
\midrule
GNAP-U & 0.9222 & $\pm$ 0.0041 & \multicolumn{2}{c}{157765} & \\
\bottomrule
\end{tabular}
\end{table}

\begin{table}[htbp]
\centering
\caption{Comparison with state-of-the-art sparse CNN on CIFAR100. Our approach outperforms most methods combining the criteria of accuracy and number of weights without the dependence on a pre-defined initial dense network.}
\label{tab:sota_cifar100}
\footnotesize
\begin{tabular}{l@{\hspace{0.5em}}r@{\hspace{0.5em}}l@{\hspace{0.5em}}r@{\hspace{0.5em}}l@{\hspace{0.5em}}c}
\toprule
Method & \multicolumn{2}{c}{Accuracy} & \multicolumn{2}{c}{Weights} & Base architecture \\
\midrule
SimpleNet~\cite{SimpleNet} & \multicolumn{2}{c}{0.6468} & \multicolumn{2}{c}{310000} &  \\
SNIP~\cite{lee2018} & 0.6554 & $\pm$ 0.0026 & \multicolumn{2}{c}{93006} & ResNet32 \\
SNIP~\cite{lee2018} & 0.2534 & $\pm$ 0.0916 & \multicolumn{2}{c}{410968} & VGG19 \\
GraSP~\cite{wang2019} & \textbf{0.6584} & $\pm$ 0.0014 & \multicolumn{2}{c}{\textbf{93006}} & ResNet32 \\
GraSP~\cite{wang2019} & 0.6584 & $\pm$ 0.0036 & \multicolumn{2}{c}{205484} & VGG19 \\
SRatio~\cite{su2020} & \textbf{0.6708} & $\pm$ 0.0041 & \multicolumn{2}{c}{\textbf{93006}} & ResNet32 \\
SRatio~\cite{su2020} & 0.6500 & $\pm$ 0.0022 & \multicolumn{2}{c}{205484} & VGG19 \\
LTH~\cite{frankle2018} & \textbf{0.6680} & $\pm$ 0.0049 & \multicolumn{2}{c}{\textbf{93006}} & ResNet32 \\
LTH~\cite{frankle2018} & 0.0980 & $\pm$ 0.1598 & \multicolumn{2}{c}{410968} & VGG19 \\
IMP~\cite{frankle2020} & 0.6467 & $\pm$ 0.0036 & \multicolumn{2}{c}{37203} & ResNet32 \\
IMP~\cite{frankle2020} & 0.0403 & $\pm$ 0.0469 & \multicolumn{2}{c}{205484} & VGG19 \\
RigL~\cite{evci2020} & 0.6446 & $\pm$ 0.0036 & \multicolumn{2}{c}{37203} & ResNet32 \\
RigL~\cite{evci2020}  & 0.6748 & $\pm$ 0.0036 & \multicolumn{2}{c}{410968} & VGG19 \\
ART~\cite{Glandorf_2023_ICCV} & \textbf{0.6586} & $\pm$ 0.0022 & \multicolumn{2}{c}{\textbf{18601}} & ResNet32 \\
ART~\cite{Glandorf_2023_ICCV} & \textbf{0.6453} & $\pm$ 0.0024 & \multicolumn{2}{c}{\textbf{41097}} & VGG19 \\
\midrule
GNAP-U & 0.6656 & $\pm$ 0.0086 & \multicolumn{2}{c}{179857} & \\
\bottomrule
\end{tabular}
\end{table}

Overall, our approach is able to obtain architectures that are very sparse with good accuracy. 
Some methods (marked in bold) obtain a comparable or slightly higher accuracy with fewer parameters. Note however, that for most of them (except ART) this depends on the base architecture. 
In contrast, our method is not based on a specific initial architecture and can grow to an appropriate size from a very small model. 
Concerning ART~\cite{Glandorf_2023_ICCV}, it uses an adaptive regularisation term for pruning which could potentially be integrated into our approach. 
Also note that some of these SOTA methods use the \emph{test set} during training in some way, which we do not.

\begin{figure}
\includegraphics[width=1.0\linewidth]{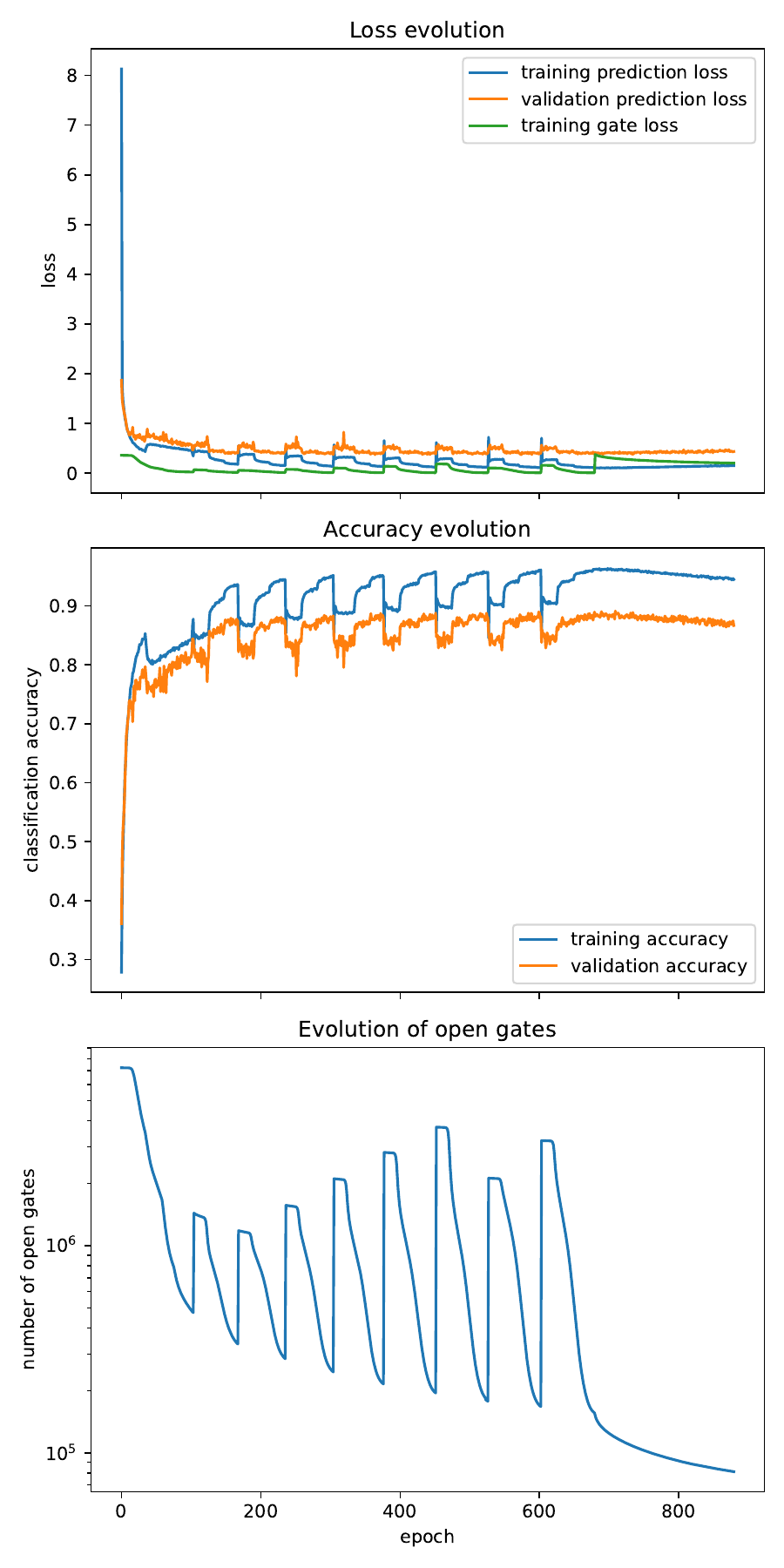}
\caption{The evolution of the different losses and number of open gates during training of our GNAP model (here GNAP-U on CIFAR10) (Please note that the scale for the number of open gates is logarithmic).}
\label{fig:losses}
\end{figure}

Figure~\ref{fig:losses} shows the evolution of accuracies, training and validation losses and of the number of open gates during training on the CIFAR10 dataset.
We can clearly see the abrupt changes when the growing phases are initiated, and the smooth evolution and recovery during the gradient descent phases. Surprisingly, it seems the growing phases actually help find smaller models. The added weights most likely help explore more effective solutions, hence why the model keeps getting smaller and smaller despite the increasingly larger growing phases.

\begin{figure}
\hspace{-3mm}\includegraphics[width=1.1\linewidth]{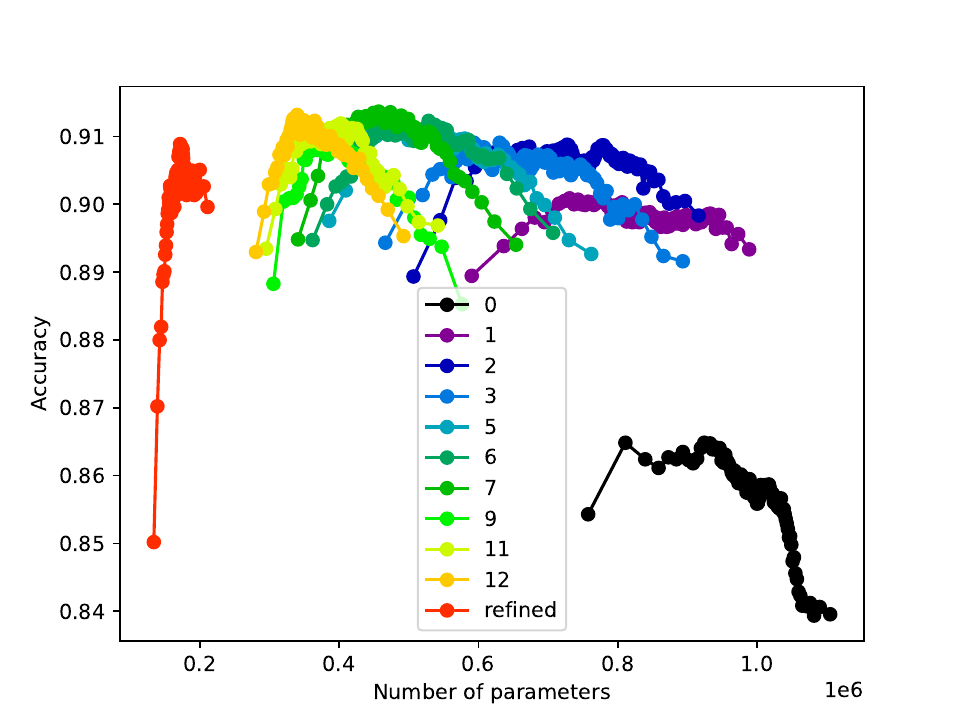}
\caption{The evolution of the accuracy vs. number of parameters. Each curve is a snapshot of the model before a given growing phase (e.g. "0" is the model before the first growing phase). The points of a curve represent different gate thresholds $\gamma$ at inference.}
\label{fig:threshold_cifar10}
\end{figure}

We conducted a more detailed analysis of the evolution of the number of active links (weights) and test accuracy of a training run on CIFAR10 with 12 growing phases. 
Figure~\ref{fig:threshold_cifar10} shows the results, where each curve represents a snapshot at a given point in time during training, i.e. before each growing phase and one after the final refinement phase. 
Each curve is made of 99 points representing different gate thresholds $\gamma$ applied at inference, from 0.01 to 0.99. Threshould 0.00 is not shown because it would display the significantly bigger unpruned network, which would make the figure unreadable. Likewise, threshold 1.00 would display the completely pruned network, with absolutely no weights left, which would similarly make the figure unreadable if shown. These curves provide us with some interesting insights. The most obvious one is that the maximum accuracy is always obtained with the trivial threshold of 0.5, showing that modulating the threshold does not necessarily lead to more interesting accuracy/complexity trade-offs of a given model. One can also notice that the first growing phases clearly increase the test accuracy up to some saturation point, and that the number of parameters is reduced with each growing phase. At the cost of a (small) loss in accuracy, the final refinement shows its abilitity to find a much light model compared to the last growing phase. Another interesting behavior is how the difference in the number of weights between threshold 0.99 and 0.01 decreases with each new curve : this means that most gate weights have either more than 99\% or less than 1\% chance of being opened for each forward pass. Effectiveley, the model seems to be converging towards a deterministic behaviour, where gate weights are either 100\% or 0\%. The refined curve also has an additional property: accuracy significantly worsens by removing only a small amount of weights. This suggests that the model has very little redundancy, and that the autonomous pruning of GNAP sucessfully kept only the most vital model weights. Finally, one can notice that weights with very high threshold at the beginning of training can nevertheless be pruned later on: curve 0, representing the model just before the first growing phase, has over 700,000 weights with a threshold above 99\%, yet the refined curbe have only slightly more than 200,000 weights with a threshold above 1\%, meaning that, at minimum, about 500,000 weights with a threshold above 99\% at the begging of training got pruned later on, and fell to a threshold below 1\%. Note that it is only a minimum, because the weights added by the growing phases have not even been taken into account ; in reality, it is very possible that many of the 200,000 weights were added by the growing phases, and therefore that the number of pruned weights is even higher. 

\begin{figure*}
\hspace{-7mm}\includegraphics[width=1.1\textwidth]{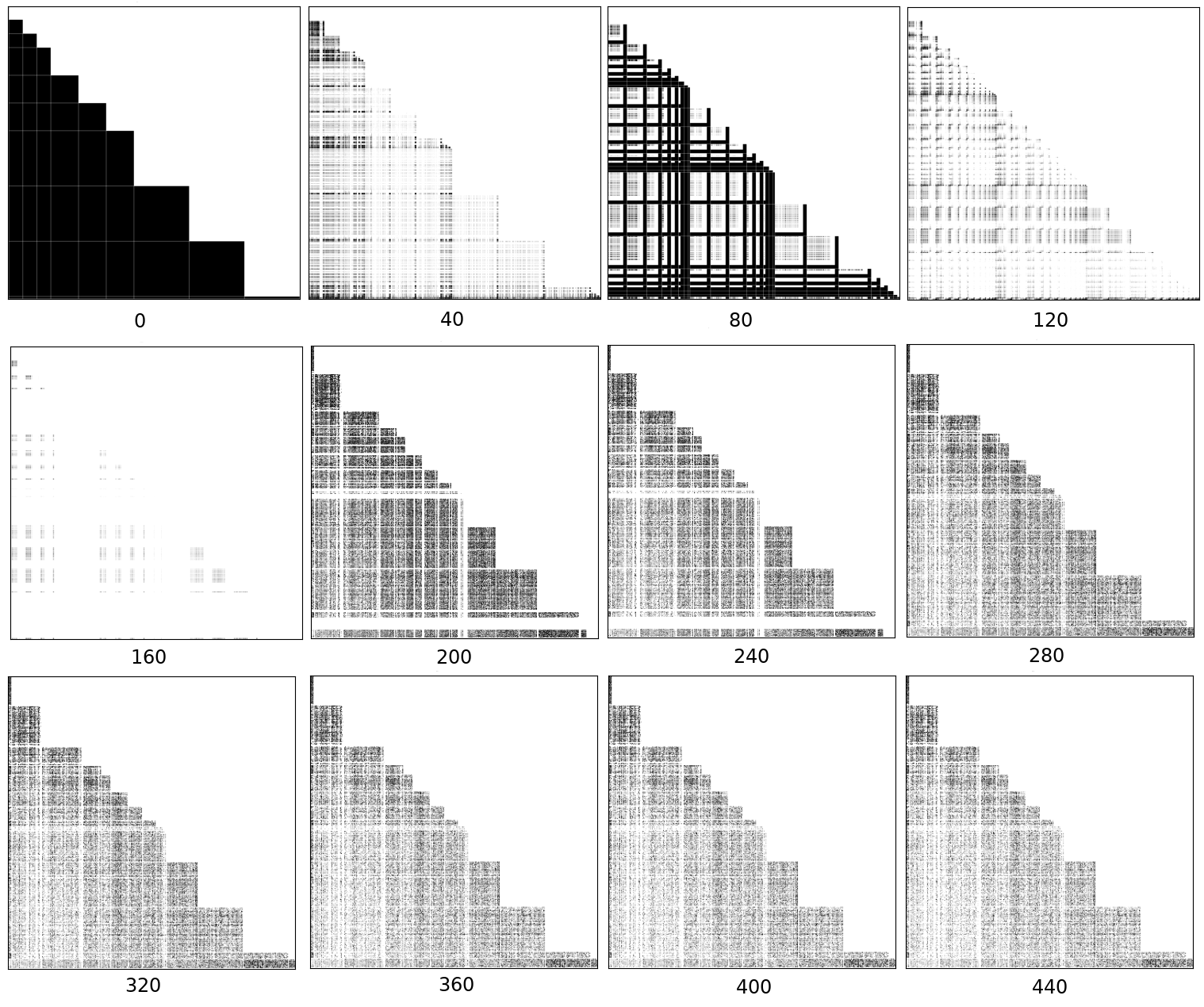}
\caption{The evolution of links (open gates) of our GNAP model during training (here GNAP-U on CIFAR10). Numbers indicate training epochs. A pixel at row i and column j represents the link going from neuron j to neuron i. The intensity of a pixel represents the proportion of open gates between two neurons/feature maps.}
\label{fig:stairs}
\end{figure*}

Finally, in Fig.~\ref{fig:stairs}, we illustrate the evolution of open gates during a typical training run (here on CIFAR10 for GNAP-U). 

Neurons that have no open gates, i.e.\ lines that are entirely blank, are removed during training (and the images in the figure are rescaled to the same size).
One can see that the algorithm very quickly closes many gates (e.g.\ at epoch 40), and additive phases add new neurons (e.g.\ at epoch 80). 
After epoch 160, many neurons are removed at once, which suddenly leads to a much darker and denser image in the figure (epoch 200).
The topology converges to a state where gates are rather uniformly opened over the network, which probably means that all remaining weights are equally useful for the model and which may be a desirable property.

\section{Conclusion}

In this paper, we introduced Growing Networks with Autonomous Pruning (GNAP) for image classification. GNAP can change their size during training, growing new parameters when needing more expressive power, and otherwise pruning their weights with complete autonomy by gradient descent, while simultaneously training for classification. In our experiments, we showed that GNAP can achieve high accuracy with extremely sparse networks, using both structured and unstructured pruning. 

In the future, we will investigate different strategies for adding neurons (when, where and how many), and how to initialise them. 
Further, the adaptation of GNAP for continual learning and non i.i.d.\ data is an interesting and promising research direction.

\section*{Acknowledgements}
This work was granted access to the HPC resources of IDRIS under the allocation AD011014045R2 made by GENCI. 

The authors acknowledge the ANR – FRANCE (French National Research Agency) for its financial support of the RADYAL project n°23-IAS3-0002.

\bibliographystyle{elsarticle-num-names} 
\bibliography{main_nn}

@STRING{IEEE_J_NNLS = "{IEEE} Transactions on Neural Networks and Learning Systems"}

@STRING{C_CVPR = "Computer Vision and Pattern Recognition (CVPR)"}

@STRING{C_ECML = "European Conference on Machine Learning (ECML)"}

@STRING{C_ICML = "International Conference on Machine Learning (ICML)"}

@STRING{C_ICLR = "International Conference on Learning Representations (ICLR)"}

@STRING{C_NeurIPS = "Advances in Neural Information Processing Systems (NeurIPS)"}

@misc{NAS,
    title={Neural Architecture Search: Insights from 1000 Papers}, 
    author={Colin White and Mahmoud Safari and Rhea Sukthanker and Binxin Ru and Thomas Elsken and Arber Zela and Debadeepta Dey and Frank Hutter},
    year={2023},
    eprint={2301.08727},
    archivePrefix={arXiv},
    primaryClass={cs.LG},
    url={https://arxiv.org/abs/2301.08727} 
}

@misc{SGDPruning,
      title={Neural Network Pruning by Gradient Descent}, 
      author={Zhang Zhang and Ruyi Tao and Jiang Zhang},
      year={2023},
      eprint={2311.12526},
      archivePrefix={arXiv},
      primaryClass={cs.LG},
      url={https://arxiv.org/abs/2311.12526}, 
}

@inproceedings{GradMax,
      title={{GradMax: Growing Neural Networks using Gradient Information}}, 
      author={Utku Evci and Bart van Merriënboer and Thomas Unterthiner and Max Vladymyrov and Fabian Pedregosa},
      year={2022},
      booktitle={ICLR}
}

@article{TINY,
      title={Growing Tiny Networks: Spotting Expressivity Bottlenecks and Fixing Them Optimally}, 
      author={Manon Verbockhaven and Sylvain Chevallier and Guillaume Charpiat and Théo Rudkiewicz},
      year={2024},
      journal = {Transactions on Machine Learning Research}
}

@article{SP-GAN,
  author={Song, Xiaoning and Chen, Yao and Feng, Zhen-Hua and Hu, Guosheng and Yu, Dong-Jun and Wu, Xiao-Jun},
  journal=IEEE_J_NNLS,
  title={SP-GAN: Self-Growing and Pruning Generative Adversarial Networks}, 
  year={2021},
  volume={32},
  number={6},
  pages={2458-2469}
}

@inproceedings{DenseNet,
  author={Huang, Gao and Liu, Zhuang and Van Der Maaten, Laurens and Weinberger, Kilian Q.},
  booktitle=C_CVPR,
  title={Densely Connected Convolutional Networks}, 
  year={2017},
  pages={2261-2269}
}

@inproceedings{GumbelSoftmax,
      title={Categorical Reparameterization with Gumbel-Softmax}, 
      author={Eric Jang and Shixiang Gu and Ben Poole},
      year={2016},
      booktitle={ICLR}
}

@misc{SimpleNet,
      title={Lets keep it simple, Using simple architectures to outperform deeper and more complex architectures}, 
      author={Seyyed Hossein Hasanpour and Mohammad Rouhani and Mohsen Fayyaz and Mohammad Sabokrou},
      year={2023},
      eprint={1608.06037},
      archivePrefix={arXiv},
      primaryClass={cs.CV},
      url={https://arxiv.org/abs/1608.06037}
}

@inproceedings{schmidt2007,
title={Fast Optimization Methods for L1 Regularization: A Comparative Study and Two New Approaches},
booktitle=C_ECML,
year={2007},
author={Mark Schmidt and Mark Schmidt and Glenn Fung and Glenn Fung and Rómer Rosales and Romer Rosales}
}

@INPROCEEDINGS{liu2017,
  author={Liu, Zhuang and Li, Jianguo and Shen, Zhiqiang and Huang, Gao and Yan, Shoumeng and Zhang, Changshui},
  booktitle={International Conference on Computer Vision (ICCV)}, 
  title={Learning Efficient Convolutional Networks through Network Slimming}, 
  year={2017},
  pages={2755-2763}
}

@ARTICLE{kumar2020,
title={Pruning filters with L1-norm and capped L1-norm for CNN compression},
year={2020},
author={Aakash Kumar and Aakash Kumar and Ali Muhammad Shaikh and Ali Muhammad Shaikh and Luyang Wang and Yun Li and Yun Li and Hazrat Bilal and Hazrat Bilal and Baocai Yin and Baoqun Yin},
journal={Applied Intelligence}
}

@inproceedings{sun2019,
author = {Xinlu Sun and Dianle Zhou and Xiaotian Pan and Zhiwei Zhong and Fei Wang},
title = {{Pruning filters with L1-norm and standard deviation for CNN compression}},
volume = {11041},
booktitle = {Eleventh International Conference on Machine Vision (ICMV)},
editor = {Antanas Verikas and Dmitry P. Nikolaev and Petia Radeva and Jianhong Zhou},
organization = {International Society for Optics and Photonics},
publisher = {SPIE},
pages = {110412J},
year = {2019}
}

@article{yang2019,
author = {Yang, Chen and Yang, Zhenghong and Khattak, Mateen and Yang, Liu and Zhang, Wenxin and Gao, Wanlin and Wang, Minjuan},
year = {2019},
month = {08},
pages = {1-1},
title = {Structured Pruning of Convolutional Neural Networks via L1 Regularization},
volume = {PP},
journal = {IEEE Access}
}

@inproceedings{mostafa2019,
title={Parameter efficient training of deep convolutional neural networks by dynamic sparse reparameterization},
year={2019},
author={Hesham Mostafa and Hesham Mostafa and Xin Wang and Xin Wang},
booktitle={International Conference on Machine Learning}
}

@article{dettmers2019,
  title={Sparse Networks from Scratch: Faster Training without Losing Performance},
  author={Tim Dettmers and Luke Zettlemoyer},
  journal={ArXiv},
  year={2019},
  volume={abs/1907.04840}
}

@article{mocanu2018,
author = {Mocanu, Decebal and Mocanu, Elena and Stone, Peter and Nguyen, Phuong and Gibescu, Madeleine and Liotta, Antonio},
year = {2018},
month = {06},
title = {Scalable training of artificial neural networks with adaptive sparse connectivity inspired by network science},
volume = {9},
journal = {Nature Communications}
}

@inproceedings{yuan2023,
author = {Yuan, Xin and Savarese, Pedro and Maire, Michael},
title = {Accelerated training via incrementally growing neural networks using variance transfer and learning rate adaptation},
year = {2023},
publisher = {Curran Associates Inc.},
address = {Red Hook, NY, USA},
booktitle = {Proceedings of the 37th International Conference on Neural Information Processing Systems},
articleno = {729},
numpages = {20},
location = {New Orleans, LA, USA},
series = {NIPS '23}
}

@inproceedings{wu2020,
title={Firefly Neural Architecture Descent: a General Approach for Growing Neural Networks},
year={2020},
author={Lemeng Wu and Lemeng Wu and Lemeng Wu and Bo Liu and Bo Liu and Peter Stone and Peter Stone and Qiang Liu and Qiang Liu and Qiang Liu and Qiang Liu},
booktitle={Neural Information Processing Systems}
}

@inproceedings{liu2019,
    title={Splitting Steepest Descent for Growing Neural Architectures}, 
    author={Qiang Liu and Lemeng Wu and Dilin Wang},
    year={2019},
    booktitle={Neural Information Processing Systems} 
}

@inproceedings{evci2020,
  title={Rigging the lottery: Making all tickets winners},
  author={Evci, Utku and Gale, Trevor and Menick, Jacob and Castro, Pablo Samuel and Elsen, Erich},
  booktitle= C_ICML,
  pages={2943--2952},
  year={2020},
  organization={PMLR}
}

@inproceedings{frankle2018,
  title={Finding sparse, trainable neural networks},
  author={Frankle, Jonathan and Carbin, Michael},
  booktitle= C_ICLR,
  year={2018}
}

@inproceedings{frankle2020,
  title={Linear mode connectivity and the lottery ticket hypothesis},
  author={Frankle, Jonathan and Dziugaite, Gintare Karolina and Roy, Daniel M. and Carbin, Michael},
  booktitle= C_ICML,
  year={2020}
}

@inproceedings{lee2018,
  title={{SNIP:} Single-shot network pruning based on connection sensitivity},
  author={Lee, Namhoon and Ajanthan, Thalaiyasingam and Torr, Philip},
  booktitle= C_ICLR,
  year={2018}
}

@inproceedings{su2020,
  title={Sanity-checking pruning methods: Random tickets can win the jackpot},
  author={Su, Jingtong and Chen, Yihang and Cai, Tianle and Wu, Tianhao and Gao, Ruiqi and Want, Liwei and Lee, Jason D.},
  booktitle= C_NeurIPS,
  year={2020}
}

@inproceedings{wang2019,
  title={Picking winning tickets before training by preserving gradient flow},
  author={Wang, Chaoqi and Zhang, Guodong and Grosse, Roger},
  booktitle= C_ICLR,
  year={2019}
}

@InProceedings{Glandorf_2023_ICCV,
    author    = {Glandorf, Patrick and Kaiser, Timo and Rosenhahn, Bodo},
    title     = {{HyperSparse Neural Networks}: Shifting Exploration to Exploitation Through Adaptive Regularization},
    booktitle = {Proceedings of the IEEE/CVF International Conference on Computer Vision (ICCV) Workshops},
    month     = {October},
    year      = {2023},
    pages     = {1234-1243}
}

@inproceedings{lin2020_mcunet,
 author = {Lin, Ji and Chen, Wei-Ming and Lin, Yujun and cohn, john and Gan, Chuang and Han, Song},
 booktitle = C_NeurIPS,
 pages = {11711--11722},
 title = {{MCUNet: Tiny Deep Learning on IoT Devices}},
 volume = {33},
 year = {2020}
}

\vfill

\end{document}